\documentclass[conference]{IEEEtran}
\IEEEoverridecommandlockouts
\usepackage{cite}
\usepackage{amsmath,amssymb,amsfonts}
\usepackage{bm}
\usepackage{graphicx}
\usepackage{textcomp}
\usepackage{xcolor}
\usepackage{algorithm}
\usepackage{algpseudocode}
\usepackage{booktabs}  
\usepackage{multirow}  
\usepackage{siunitx}   
\usepackage{caption}   
\usepackage{enumitem}
\usepackage{geometry}
\usepackage{hyperref} 
\usepackage{xcolor}   
\hypersetup{
    colorlinks=true, 
    linkcolor=blue, 
    citecolor=blue, 
    urlcolor=blue,  
}

\usepackage{makecell}
\usepackage{pifont}
\usepackage{tabularx}

\newcounter{gcsalgo}


\begin{document}

\title{
GCS-Bridging: Restoring Connectivity of Disconnected Convex Sets for Graph-of-Convex-Sets Motion Planning
\thanks{
This work was supported in part by the National Natural Science Foundation of China through the Basic Science Center Program for Space Robot Intelligent Manipulation under Grant T2388101, and in part by the National Natural Science Foundation of China under Grant 52475012.

1: State Key Laboratory of Robotics and Systems, Harbin Institute of Technology, Harbin 150080 Heilongjiang Province, China. 

2: School of Mechanical and Aerospace Engineering, Nanyang Technological University, Singapore 639798, Singapore

*: Corresponding Author}
\thanks{E-mail:\tt\footnotesize 
zhouxiaokai@stu.hit.edu.cn, 
cbs@hit.edu.cn, 
liuyanghit@hit.edu.cn, 
sunkui@hit.edu.cn,
boyu.ma@ntu.edu.sg, 
21b908032@stu.hit.edu.cn, 
xiezongwu@hit.edu.cn}
}

\author{Xiaokai~Zhou\textsuperscript{1},
        Baoshi~Cao\textsuperscript{1,*},
        Yang~Liu\textsuperscript{1},
        Kui~Sun\textsuperscript{1},
        Boyu~Ma\textsuperscript{2},
        Zhengpu~Wang\textsuperscript{1},
        and~Zongwu~Xie\textsuperscript{1}}

\maketitle

\begin{abstract}
Graph-of-Convex-Sets (GCS)-based trajectory optimization represents collision-free regions in configuration space as a finite collection of convex sets and directly performs collision-free trajectory planning over these sets, substantially simplifying the planning process. However, existing GCS-based trajectory planning methods generally assume sufficient connectivity among the convex regions and do not explicitly address cases in which the start and goal regions belong to different connected components of the initial GCS map. To address this limitation, we propose GCS-Bridging, which reconnects disconnected convex regions through collision-free point paths followed by convex region inflation, thereby recovering the feasibility of otherwise disconnected GCS planning problems. Extensive simulations across multiple IRIS-related algorithms and scenarios demonstrate that GCS-Bridging restores missing start-to-goal connectivity in the initial GCS map with a 99.8\% success rate. In addition, a hardware experiment on a single-arm Franka platform in a real-world scenario with initially disconnected start and goal regions validates the effectiveness of the proposed method in practical motion planning. Project website: 
\url{https://zhouxk1997.github.io/GCS_Bridging/}.
\end{abstract}


\section{Introduction}
\label{Section I}
Motion planning for redundant manipulators remains a major research topic in robotics, yet many challenges persist. A fundamental challenge arises from the nonlinear mapping between the workspace, where task-related control points are defined in the physical world, and the configuration space (hereafter referred to as C-space), where robot motions are represented by joint configurations. This issue is particularly significant in obstacle avoidance. Ideally, collision avoidance can be formulated as a simple problem directly in C-space \cite{lozano1983spatial}, substantially simplifying its mathematical representation. However, even simple obstacles in workspace can induce highly complex representations in C-space. For example, a spherical obstacle may correspond to a high-dimensional set with periodicity, nonconvexity, and curved boundaries. This complexity fundamentally arises from the highly nonlinear mapping between robot configurations and task space, which is inherent to serial kinematic structures and multi-joint coupling. It therefore constitutes one of the fundamental challenges in collision-free motion planning for redundant manipulators \cite{chen2026cssdf}, \cite{li2024configuration}.

The complex representation of obstacles in C-space has not hindered the development of numerous C-space motion planning algorithms. An intuitive strategy is to plan directly around obstacles in C-space without explicitly converting workspace obstacles into high-dimensional configuration-space obstacle geometry. Instead, collision costs are evaluated as needed during planning to directly generate collision-free trajectories. Such methods avoid explicitly representing the complete configuration-space obstacle set. This paper collectively refers to them as direct motion-generation methods because they produce a single collision-free trajectory. Representative approaches include sampling-based method \cite{thomason2024motions}, \cite{wilson2025nearest}, \cite{huang2025prrtc}, optimization-based method \cite{sundaralingam2023curobo}, \cite{huang2024diffusionseeder} \cite{tam2024bomp}, and learning-based method \cite{dalal2024neural}, \cite{zhang2024robotdiffuse}. These methods primarily generate feasible motions for a given start-to-goal planning query rather than first constructing a reusable representation of collision-free space in C-space. However, the collision information obtained during a single planning process is tightly coupled with the current trajectory generation. It is therefore difficult to convert this information into an explicit representation of collision-free space in C-space. Even after a collision-free trajectory is obtained, the resulting planning information generally cannot be directly reused for new planning queries or global trajectory optimization.

\begin{figure*}[t]
    \centering
    \includegraphics[width=0.97\textwidth]{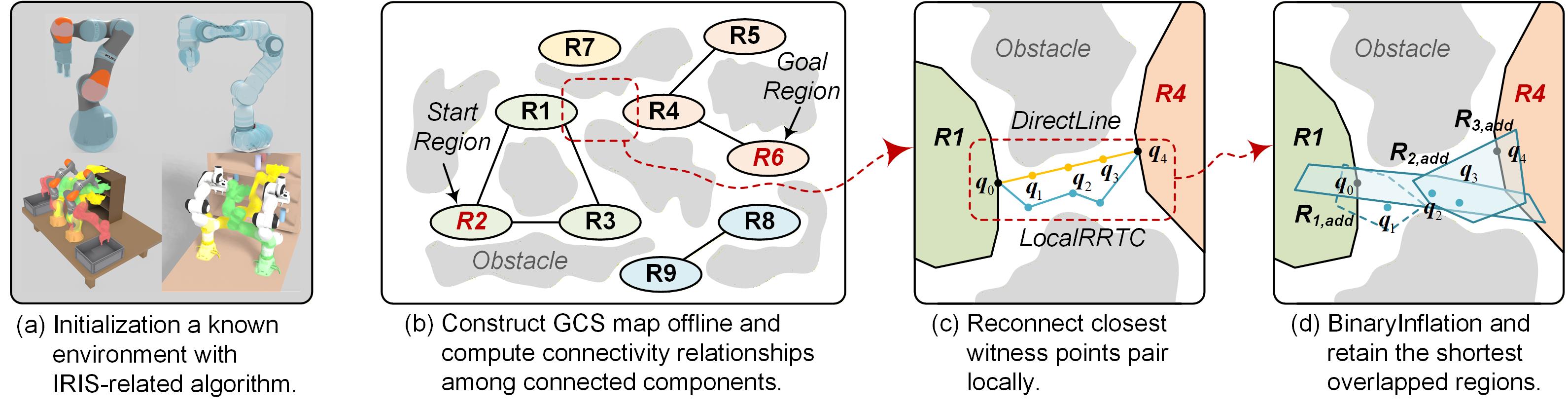}
    \caption{A conceptual illustration of GCS-Bridging in an abstract configuration space. (a) Initialization is first performed in a fully known environment using multiple IRIS-related algorithms. (b) An offline GCS map is then constructed, in which the regions containing the start and goal configurations may be disconnected. The connected components and the relationships among all regions are subsequently computed from the GCS map. (c) Based on the global connectivity information of the GCS map, the algorithm attempts to locally reconnect the closest regions pair while ensuring that the entire reconnection path remains collision-free. (d) BinaryInflation is then performed along the path; once the two disconnected regions are reconnected, only the shortest connection is retained.}
    \label{Figure 1: Overview of GCS-Bridging}
\end{figure*}

In contrast, although workspace obstacles are difficult to represent explicitly in C-space, many methods enable explicit collision-free planning by approximating collision-free configuration sets around obstacles. Representative examples include the union-of-balls representation used by Yang and LaValle \cite{yang2004sampling} and the polyhedral regions constructed by Deits and Tedrake \cite{deits2015computing}. Building on these concepts, the Graph-of-Convex-Sets trajectory optimization (GCSTrajOpt) method \cite{marcucci2023motion} was developed using convex regions generated by Iterative Regional Inflation by Semidefinite programming (IRIS). GCSTrajOpt represents collision-free configuration sets in C-space as convex regions satisfying $\boldsymbol{A} q \le \boldsymbol{b}$ and performs global convex optimization over connected collision-free regions, providing a concise and efficient framework for collision-free planning. Several related planning methods \cite{werner2024faster}, \cite{werner2025superfast}, \cite{clark2025plan}, \cite{natarajan2024implicit} have subsequently been developed based on the same principle of generating connected convex regions in joint space before trajectory optimization. However, existing GCS-based motion-planning pipelines generally focus on trajectory optimization over a preconstructed graph of collision-free convex regions, while the recovery of missing start-to-goal connectivity when the GCS map is internally disconnected has received considerably less attention.

In this work, we further extend the IRIS-based framework for collision-free convex region generation in configuration space and GCS-based motion planning, with particular emphasis on GCSTrajOpt infeasibility caused by insufficient convex-region coverage and missing region connectivity. To address this issue, we propose GCS-Bridging. Our contributions are summarized as follows:

(1) We formulate insufficient connectivity in the initial GCS map as a connectivity repair problem;

(2) We propose a GCS map reconnection algorithm based on cross-connected-component, locally adaptive discrete-path reconnection and connectivity-aware region inflation;

(3) We validate the proposed method using four IRIS algorithms, 6 randomized single- and dual-arm simulation scenarios, and real-robot experiment, achieving an overall reconnection success rate of 99.8\% with RRT-C-based method, and 100\% with the hybrid method.

The remainder of this paper is structured as follows. Section \ref{Section II} formulates the problem theoretically. Section \ref{Section III} presents the proposed GCS-Bridging algorithm in detail, including locally adaptive discrete-path reconnection and connectivity-aware region inflation. Section \ref{Section IV} reports the experimental results. Finally, Section \ref{Section V} discusses the conclusions and limitations.

\section{Problem Formulation}
\label{Section II}
Consider a redundant manipulator with configuration $\boldsymbol{q} \in \boldsymbol{C} \subseteq \mathbb{R}^{n}$, where $\boldsymbol{C}$ denotes the C-space of the manipulator. Let $\boldsymbol{C}_{\text{free}} \subseteq \boldsymbol{C}$ denote the collision-free configuration space, and let $\boldsymbol{q}_{s}$ and $\boldsymbol{q}_{g}$ denote the start and goal configurations, respectively.

Collision-free convex regions can be generated in C-space using heuristic or probabilistically complete region inflation methods. Assume that each generated convex region satisfies $\boldsymbol{R}_{i} \subseteq \boldsymbol{C}_{\text{free}}$. Given a finite collection of collision-free convex regions:
\begin{equation}
    \label{Equation 1: Convex Regions Set}
    \boldsymbol{R} = \{ \boldsymbol{R}_{1}, \boldsymbol{R}_{2}, ..., \boldsymbol{R}_{n_r} \}
\end{equation}
a GCS map with connectivity information is constructed as
\begin{equation}
    \label{Equation 2: GCS map}
    \boldsymbol{G} = \left( \boldsymbol{V}, \boldsymbol{E} \right)
\end{equation}
where
\begin{equation}
    \label{Equation 3: GCS map components}
\begin{gathered}
    \boldsymbol{V} = \{ \boldsymbol{v}_{1}, \boldsymbol{v}_{2}, ..., \boldsymbol{v}_{n_c} \} \\
    \boldsymbol{E} = \{ \boldsymbol{e}_{1}, \boldsymbol{e}_{2}, ..., \boldsymbol{e}_{n_c} \}
\end{gathered}    
\end{equation}

Here, $\boldsymbol{V}$ is the set of connected components, where each $\boldsymbol{v}_{i}$ contains a subset of mutually connected convex regions. For any two regions $\boldsymbol{R}_{j},\boldsymbol{R}_{k}\in\boldsymbol{v}_{i}$, there exists at least one feasible sequence of connected convex regions linking $\boldsymbol{R}_{j}$ and $\boldsymbol{R}_{k}$. And $\boldsymbol{E}$ denotes the connectivity information of all connected components, where $\boldsymbol{e}_{i}$ contains all connectivity edges among the convex regions within $\boldsymbol{v}_{i}$. $\boldsymbol{e}_{i}$ can be expressed as following equation:
\begin{equation}
    \label{Equation 4: Edge definition}
    \boldsymbol{e}_{i}=\left\{(\boldsymbol{R}_{j},\boldsymbol{R}_{k})\mid \boldsymbol{R}_{j},\boldsymbol{R}_{k}\in\boldsymbol{v}_{i}, \;\boldsymbol{R}_{j}\cap\boldsymbol{R}_{k}\neq\varnothing \right\}
\end{equation}
$\boldsymbol{e}_{i}$ is the core of the GCSTrajOpt solving process, and optimization can only be achieved with the edge information of adjacent regions. Using the GCSTrajOpt solver, for any GCS trajectory optimization problem:
\begin{equation}
\begin{aligned}
    \label{Equation 5: Optimization problem}
    \underset{\boldsymbol{q},T}{\operatorname{minimize}} \quad 
    & aL(\boldsymbol{q}) + bT \\
    \text{subject to} \quad 
    & \boldsymbol{q}(t) \in \boldsymbol{R}_1 \cup \cdots \cup \boldsymbol{R}_{n_{r}},
    && \forall t \in [0,T], \\
    & \boldsymbol{q}(0) = \boldsymbol{q}_s, \quad \boldsymbol{q}(T) = \boldsymbol{q}_g, \\
\end{aligned}
\end{equation}

This problem may be infeasible because the convex regions $\boldsymbol{R}_{s}$ and $\boldsymbol{R}_{g}$ containing $\boldsymbol{q}_s$ and $\boldsymbol{q}_g$, respectively, do not belong to the same connected component $\boldsymbol{v}_{i}$. Consequently, the GCSTrajOpt solver cannot identify a connected sequence of convex regions between $\boldsymbol{q}_s$ and $\boldsymbol{q}_g$, and thus cannot generate a collision-free trajectory in C-space.

The objective of GCS-Bridging is therefore to construct additional collision-free convex regions that connect the disconnected components containing the start and goal regions, thereby restoring a feasible connected region sequence for subsequent GCS trajectory optimization. And, whenever possible, reduce the C-space to a single connected component to enlarge the set of convex-region sequences available to GCSTrajOpt.

\section{GCS-Bridging Algorithm}
\label{Section III}

In this section, the design and implementation of the GCS-Bridging algorithm, illustrated in Fig.\ref{Figure 1: Overview of GCS-Bridging} and Algorithm~\ref{Algorithm 1: GCS-Bridging Algorithm}, are presented. In general, GCS-Bridging first computes the closest point pair between two convex regions in configuration space and establishes an initial bridge using $\mathrm{LocalRRTC}$. Convex region connectivity is then constructed by inflating the RRT-C path. The key contribution of GCS-Bridging is that it no longer relies solely on a predefined convex-region connectivity GCS map, but actively detects and repairs missing connectivity in the graph topology.

More specifically, the connectivity graph $\boldsymbol{G}$ and connected components information $\boldsymbol{V}$ are first constructed from the initially generated regions set $\boldsymbol{R}$, corresponding to lines 3-4 of Algorithm~\ref{Algorithm 1: GCS-Bridging Algorithm}. If $\boldsymbol{V}$ contains only one component, all regions are fully connected and no further bridging is required. Otherwise, the Euclidean distances in C-space between convex regions belonging to different connected components are computed, together with the corresponding closest witness points pair, and the candidate pairs are sorted in ascending order of distance, corresponding to line 8 of Algorithm~\ref{Algorithm 1: GCS-Bridging Algorithm}. After sorting, each pair is connected using either $\mathrm{LocalRRTC}$ or $\mathrm{DirectLine}$ followed by $\mathrm{LocalRRTC}$, corresponding to line 10 of Algorithm~\ref{Algorithm 1: GCS-Bridging Algorithm}. If a connection is found, $\mathrm{BinaryInflation}$ is applied by selecting points along the path for IRIS inflation and evaluating the connectivity of the resulting regions, corresponding to lines 16-24 of Algorithm~\ref{Algorithm 1: GCS-Bridging Algorithm}. If the two points cannot be connected or connected regions cannot be generated, the corresponding pair is recorded in $\mathrm{FailedPairs}$ and excluded from further connection attempts. Finally, the GCS map reconnected by GCS-Bridging is returned.

\subsection{LocalRRTC Bridging}
\label{Section III.A}

In lower-dimensional scenarios, such as the joint space of a 7-DoF redundant manipulator, DirectLine connection can be highly efficient. However, in complex environments, connecting two collision-free configurations requires the entire line segment between them to remain collision-free, resulting in a lower success rate. One major challenge in joint-space planning is the curse of dimensionality. In high-dimensional C-space, planning efficiency can degrade sharply, computational cost can grow exponentially, and exploration regions may collapse, preventing effective obstacle avoidance. These issues pose severe challenges for practical applications. RRT-Connect typically performs bidirectional exploration in C-space and is well suited for planning between arbitrary start and goal configurations. In GCS-Bridging, however, the objects to be connected are not two isolated configurations, but two previously generated collision-free convex regions. Therefore, RRT-C serves as a local bridge between two disconnected convex regions and searches only for a feasible connection between the corresponding components. The corresponding algorithm is shown in Algorithm~\ref{Algorithm 2: LocalRRTC}.

First, the minimum-distance witness points lie on the boundaries of the convex regions. However, floating-point error, algorithmic error, or collision-detection error may cause these witness points to fail collision checks, leading RRT-C to start from invalid endpoints and generate unreliable paths. Therefore, $\mathrm{ValidateWitnessPoints}$ first performs collision checks on both endpoints. For any colliding witness point, the seed point $\boldsymbol{s}_{i}$ used to generate the corresponding convex region $\boldsymbol{R}_{i}$ is employed to retract the point by a finite distance toward the interior of the region, after which the candidate point is rechecked, corresponding to Algorithm~\ref{Algorithm 2: LocalRRTC} line 1-4:
\begin{equation}
    \label{Equation 6: Repair witness points}
    \boldsymbol{q}^{*} = \boldsymbol{q} + \lambda \frac{\boldsymbol{s}_i-\boldsymbol{q}}{\|\boldsymbol{s}_i-\boldsymbol{q}\|_2}
\end{equation}

After all candidate points satisfy the joint-limit, original-region membership, and collision-free constraints, $\mathrm{LocalRRTC}$ bridging is performed within an axis-aligned local region to mitigate the curse of dimensionality in C-space. The local margin of this region along each axis is defined as follows:
\begin{equation}
    \label{Equation 7: Local sampling margin}
    \rho_{l} = \min\Big(0.1Q_l,\ \max\big(d_{ab},\ 0.5|q_{a,l} - q_{b,l}|,\ 0.005Q_l\big)\Big)
\end{equation}

\newsavebox{\leftalgbox}

\begin{figure*}[t]
\centering

\begin{lrbox}{\leftalgbox}
\begin{minipage}[t]{0.485\textwidth}
\vspace{0pt}   

\hrule
\vspace{2pt}
\refstepcounter{gcsalgo}
\textbf{Algorithm \thegcsalgo} GCS-Bridging Algorithm
\label{Algorithm 1: GCS-Bridging Algorithm}
\vspace{2pt}
\hrule
\vspace{3pt}

\begin{algorithmic}[1]
    \State \textbf{Input}:$\boldsymbol{R}=\{\boldsymbol{R}_1,\ldots,\boldsymbol{R}_n\}$,$\boldsymbol{Q}_{\min}$,$\boldsymbol{Q}_{\max}$
    \State $\mathrm{FailedPairs}\gets\varnothing$
    \State $\boldsymbol{G}\gets \operatorname{BuildConnectivityGraph}(\boldsymbol{R})$
    \State $\boldsymbol{V}\gets\operatorname{ConnectedComponents}(\boldsymbol{G})$
    \If{$|\boldsymbol{V}|=1$}
    \State \Return $\boldsymbol{R}$
    \EndIf
    \State $\boldsymbol{L}\gets \operatorname{RankCrossComponentPairs}(\boldsymbol{R},\boldsymbol{V},\mathrm{FailedPairs})$
    \ForAll{$(\boldsymbol{R}_a,\boldsymbol{R}_b,\boldsymbol{q}_a,\boldsymbol{q}_b)\in\boldsymbol{L}$}
        \State $\boldsymbol{P}\gets
        \operatorname{LocalRRTC}
        (\boldsymbol{q}_a,\boldsymbol{q}_b,\boldsymbol{R}_a,\boldsymbol{R}_b)\text{ or}$
        \Statex \hspace{\algorithmicindent}
        $\phantom{\boldsymbol{P}\gets}\operatorname{DirectLine} \& \operatorname{LocalRRTC}
        (\boldsymbol{q}_a,\boldsymbol{q}_b,\boldsymbol{R}_a,\boldsymbol{R}_b)$ 
        \If{$\boldsymbol{P}=\varnothing$}
            \State $\mathrm{FailedPairs}\gets \mathrm{FailedPairs}\cup {(\boldsymbol{R}_a,\boldsymbol{R}_b)}$
            \State \textbf{continue}
        \EndIf
        \State $(\boldsymbol{R}_{P},\mathrm{Connected})\gets \mathrm{BinaryInflation}(\boldsymbol{P},\boldsymbol{R}_a,\boldsymbol{R}_b)$
        \If{$\mathrm{Connected}$}
            \State $\boldsymbol{R}\gets \boldsymbol{R}\cup\boldsymbol{R}_{P}$
            \State Record $(\boldsymbol{R}_a,\boldsymbol{R}_b)$ as bridged
            \State $\boldsymbol{G}\gets \operatorname{BuildConnectivityGraph}(\boldsymbol{R})$
            \State $\boldsymbol{V}\gets\operatorname{ConnectedComponents}(\boldsymbol{G})$
            \State \textbf{continue}
        \Else
            \State $\mathrm{FailedPairs}\gets \mathrm{FailedPairs}\cup {(\boldsymbol{R}_a,\boldsymbol{R}_b)}$
            \EndIf
    \EndFor
    \State \Return $\boldsymbol{R}$
\end{algorithmic}

\vspace{3pt}
\hrule

\end{minipage}
\end{lrbox}

\begin{minipage}[t]{0.485\textwidth}
\vspace{0pt}   
\usebox{\leftalgbox}
\end{minipage}
\hfill
%
\begin{minipage}[t][\dimexpr\ht\leftalgbox+\dp\leftalgbox\relax][t]
{0.485\textwidth}
\vspace{0pt}   

\hrule
\vspace{2pt}
\refstepcounter{gcsalgo}
\textbf{Algorithm \thegcsalgo} $\mathrm{LocalRRTC}(\boldsymbol{q}_a,\boldsymbol{q}_b,\boldsymbol{R}_a,\boldsymbol{R}_b)$
\label{Algorithm 2: LocalRRTC}
\vspace{1pt}
\hrule
\vspace{3pt}

\begin{algorithmic}[1]
    \State $(\boldsymbol{q}_a,\boldsymbol{q}_b)\gets
    \mathrm{ValidateWitnessPoints}(\boldsymbol{q}_a,\boldsymbol{q}_b,\boldsymbol{R}_a,\boldsymbol{R}_b)$
    \If{$\boldsymbol{q}_a=\varnothing$ \textbf{or} $\boldsymbol{q}_b=\varnothing$}
    \State \Return $\varnothing$
    \EndIf
    \State $\boldsymbol{\rho} \gets \mathrm{InitialLocalMargin}
    \left(\boldsymbol{q}_a, \boldsymbol{q}_b, \boldsymbol{Q}_{\min}, \boldsymbol{Q}_{\max}\right)$
    \For{$k = 0,\ldots,N_{\text{maxiter}}-1$}
        \State $\boldsymbol{B}_k \gets \boldsymbol{B}\left(2^k\boldsymbol{\rho}\right)$
        \For{$r = 1,\ldots,N_{\text{restart}}$}
            \State $\boldsymbol{P} \gets \mathrm{RRTConnect}\left( \boldsymbol{q}_a, \boldsymbol{q}_b, \boldsymbol{B}_k \right)$
            \If{$\boldsymbol{P} \neq \varnothing$ \textbf{and} $\mathrm{DenseCollisionCheck}(\boldsymbol{P})$}
            \State \Return $\boldsymbol{P}$
            \EndIf
        \EndFor
    \EndFor
\end{algorithmic}

\vspace{3pt}
\hrule

\vfill

\hrule
\vspace{2pt}
\refstepcounter{gcsalgo}
\textbf{Algorithm \thegcsalgo} $\mathrm{BinaryInflation}(\boldsymbol{P},\boldsymbol{R}_a,\boldsymbol{R}_b)$
\label{Algorithm 3: BinaryInflation}
\vspace{1pt}
\hrule
\vspace{3pt}
\begin{algorithmic}[1]
    \State $\boldsymbol{O} \gets \mathrm{BinaryMidpointOrder}(\boldsymbol{P})$
    \State $\boldsymbol{R}_{\text{staged}} \gets \varnothing$
    \For{$\mathcal{M} \in \mathrm{ConsecutiveBatches}(\boldsymbol{O})$}
        \State $\boldsymbol{R}_{\text{new}} \gets \mathrm{ParallelInflation}\left( \mathcal{M} \right)$
        \State $\boldsymbol{R}_{\text{staged}} \gets \boldsymbol{R}_{\text{staged}} \cup \boldsymbol{R}_{\text{new}}$
        \State $\boldsymbol{G}_{\text{tmp}} \gets \mathrm{BuildConnectivityGraph} \left(\boldsymbol{R} \cup \boldsymbol{R}_{\text{staged}}\right)$
        \If{$\operatorname{ConnectivityCheck}(\boldsymbol{G}_{\text{tmp}},\boldsymbol{R}_a,\boldsymbol{R}_b)$}
            \State $\boldsymbol{L}_{s} \gets \mathrm{ShortestOverlapPath} \left(\boldsymbol{G}_{\text{tmp}}, \boldsymbol{R}_a, \boldsymbol{R}_b\right)$
            \State \Return $\left(\mathrm{True}, \boldsymbol{L}_{s} \cap \boldsymbol{R}_{\text{staged}}\right)$
        \EndIf
    \EndFor
\end{algorithmic}

\vspace{3pt}
\hrule

\end{minipage}

\end{figure*}

Here, $\rho_{l}$ denotes the half local margin along the $l$-th dimension, $d_{ab}=\left \|\boldsymbol{q}_a-\boldsymbol{q}_b\right \|_2$ represents the distance between the start and goal configurations of $\mathrm{LocalRRTC}$ bridging, and $Q_l = \left| Q_{\text{max},l} - Q_{\text{min},l} \right|$ denotes the total joint range of the $l$-th dimension. The resulting axis-aligned local region is denoted as $\boldsymbol{B}\left( \rho \right)$ and defined as follows:
\begin{align}
    \label{Equation 8: Local region definition}
    \nonumber \boldsymbol{B}\left( \rho \right) = \prod_{l=1}^{n} \Bigl[ & \max\left(Q_{\text{min},l},\min(q_{a,l},q_{b,l})-\rho_l\right), \\ 
    & \min\left(Q_{\text{max},l},\max(q_{a,l},q_{b,l})+\rho_l\right) \Bigr]
\end{align}

$\mathrm{LocalRRTC}$ bridging is performed within the axis-aligned local region defined by equation (\ref{Equation 8: Local region definition}). This search domain restricts RRT-C to the vicinity of the witness points while preserving sufficient range along each dimension to establish an initial bridge between the two regions. To improve the success rate of local bridging between disconnected convex regions in practical applications, two modifications are introduced in line 8-13 of Algorithm~\ref{Algorithm 2: LocalRRTC}: 

(1) if bridging fails within the current sampling region $\boldsymbol{B}(\boldsymbol{\rho})$, the half local margin is expanded as $\boldsymbol{\rho} = 2\boldsymbol{\rho}$ and sampling is repeated; 

(2) multiple random seeds are used during sampling to enhance exploration diversity. 

These strategies enable local sampling to establish initial connections between unconnected regions in high-dimensional configuration space.

\subsection{DirectLine Bridging}
\label{Section III.B}
In principle, $\mathrm{DirectLine}$ can efficiently connect simple regions pairs, making it well suited to lower-dimensional scenarios while remaining effective for simple cases in higher-dimensional spaces. Therefore, sequentially applying $\mathrm{DirectLine}$ followed by $\mathrm{LocalRRTC}$ is introduced as an optional strategy to improve the adaptability of GCS-Bridging across different scenarios. $\mathrm{DirectLine}$ is straightforward: continuous collision checking is performed along the witness points segment between a regions pair, and the line is accepted if no collision is detected. $\mathrm{BinaryInflation}$ is then applied to generate the corresponding convex-region connection.

\subsection{Binary Inflation}
\label{Section III.C}

The two unconnected regions are initially bridged only by the piecewise-linear path generated by RRT-C. However, GCS planning requires optimization over connected convex regions representable as $\boldsymbol{A} \boldsymbol{x} \le \boldsymbol{b}$. Therefore, the path obtained from $\mathrm{LocalRRTC}$ bridging must be further inflated to satisfy the requirements of the GCS planner, as shown in Algorithm~\ref{Algorithm 3: BinaryInflation}. The path returned by RRT-C is denoted as:

\begin{equation}
    \label{Equation 9: RRT-C path}
    \boldsymbol{P} = \left( \boldsymbol{q}_{0}, \boldsymbol{q}_{1}, ..., \boldsymbol{q}_{m-1} \right)
\end{equation}
here, $\boldsymbol{q}_{0} \in \boldsymbol{R}_{a}$ and $\boldsymbol{q}_{m-1} \in \boldsymbol{R}_{b}$, while $\boldsymbol{R}_{a}$ and $\boldsymbol{R}_{b}$ belong to different connected components. A straightforward strategy is to perform convex region inflation at every point along the path $\boldsymbol{P}$. However, this is computationally expensive, and regions generated from adjacent path points may overlap substantially, introducing redundant connected edges and increasing the scale of GCSTrajOpt. Therefore, a binary inflation strategy is adopted, performing inflation from coarse to fine starting from the midpoint, with actual connectivity of the GCS map used as the termination criterion.

\begin{table*}[h]
    \centering
    \caption{Comparison of GCS-Bridging using different IRIS inflation algorithms in the same dual-arm planning scenario.}
    \label{Table: Multiple IRIS Algorithm Comparison}
    \begin{tabular}{c
    c
    c
    c
    c
    c
    c}
    \toprule
        \makecell[c]{\textbf{Region} \\ \textbf{Generator}} 
        & 
        \makecell[c]{$N^{\text{init}}_{\boldsymbol{R}} \to N^{\text{final}}_{\boldsymbol{R}}$} 
        & 
        \makecell[c]{$N^{\text{init}}_{\boldsymbol{V}} \to N^{\text{final}}_{\boldsymbol{V}}$} 
        & 
        \makecell[c]{$T_{\text{inflate}}$ \textbf{(min)}}
        & 
        \makecell[c]{\textbf{GCS-Bridging} \\ \textbf{Iteration Times}\\ \text{[O,R,I]}} 
        & 
        \makecell[c]{\textbf{Total} $T_{\text{bridge}}$ \textbf{(min)} \\ \text{[R+I+C]}}
        & 
        \makecell[c]{\textbf{GCSTrajOpt} \\ \textbf{Cost \& Result}}
        \\
        
        \midrule
        \makecell[c]{IRIS-NP\cite{marcucci2023motion}} 
        & 
        24 $\to$ 30
        & 
        8 $\to$ 1
        & 
        507.70
        & 
        5,14,5
        & 
        3.13+159.85+0.92 
        & 
        23.94, \href{https://zhouxk1997.github.io/GCS_Bridging/result/GCS_Bridging_IRIS_Comparison_IRIS_NP.html}{click here}
        \\ 
        
        \makecell[c]{IRIS-NP2\cite{werner2024faster}} 
        & 
        24 $\to$ 28 
        & 
        \textbf{6 $\to$ 1}
        & 
        180.35 
        & 
        \textbf{4,7,4}
        & 
        \textbf{0.73+106.40+0.20}
        & 
        \textbf{19.26}, \href{https://zhouxk1997.github.io/GCS_Bridging/result/GCS_Bridging_IRIS_Comparison_IRIS_NP2.html}{click here}
        \\
        
        \makecell[c]{IRIS-ZO\cite{werner2024faster}} 
        & 
        25 $\to$ 87
        & 
        17 $\to$ 1
        & 
        20.82 
        & 
        16,102,28 
        & 
        35.65+110.19+83.53
        & 
        34.29, \href{https://zhouxk1997.github.io/GCS_Bridging/result/GCS_Bridging_IRIS_Comparison_IRIS_ZO.html}{click here}
        \\
        
        \makecell[c]{IRIS-ZO-CUDA} 
        & 
        25 $\to$ 90
        & 
        18 $\to$ 1 
        & 
        \textbf{2.28} 
        & 
        17,57,28
        & 
        22.75+24.91+116.39
        & 
        31.34, \href{https://zhouxk1997.github.io/GCS_Bridging/result/GCS_Bridging_IRIS_Comparison_IRIS_ZO_CUDA.html}{click here}
        \\
        
    \bottomrule
    \end{tabular}

    \vspace{1mm}
    \begin{minipage}{0.98\textwidth}
        \footnotesize{Column 1 specifies the IRIS algorithm. Column 2 reports the change in the number of regions from the initial state to the completion of GCS-Bridging. Column 3 reports the corresponding reduction in the number of connected components. Column 4 gives the total time required for the initial inflations. Column 5 reports the GCS-Bridging iteration counts, where O, R, and I denote the Outer-loop, RRT-C, and Inflation iteration counts, respectively. Column 6 reports the total GCS-Bridging computation time, where R+I+C denotes the time spent on RRT-C, Inflation, and Connectivity-Check, respectively. Column 7 represents the cost and visualized result of trajectory generated by GCSTrajOpt Algorithm. Notice, IRIS-based algorithms do not provide an absolute collision-free guarantee; therefore, trajectories generated by GCSTrajOpt may still contain collisions. Such collisions can be mitigated through subsequent trajectory optimization. However, this issue is beyond the scope of this work and is therefore not further addressed.}
    \end{minipage}
\end{table*}

Before inflation, within the path-index interval $\left[ 0, m-1 \right]$ returned by $\mathrm{LocalRRTC}$, the interval $\left[ u, v \right]$ with the largest span is selected, and its midpoint index $t = \lfloor {u+v}/{2}  \rfloor $ is computed. The IRIS inflation algorithm is then applied at this path point to generate a collision-free convex region. The interval is subsequently divided into $\left[ u, t \right]$ and $\left[ t, v \right]$, and the longer interval is processed next. Here, binary refers only to the order of inflation and not to a search over collision boundaries. After each inflation, the GCS map in equation (\ref{Equation 2: GCS map}) is recomputed to determine whether the endpoint regions $\boldsymbol{R}_{a}$ and $\boldsymbol{R}_{b}$ belong to the same connected component. This criterion avoids fully reproducing the RRT path or generating a convex region at every internal path point. The final requirement is the existence of a connected chain of convex regions:
\begin{equation}
    \label{Equation 10: Connected convex regions}
    \boldsymbol{R}_{a} \leftrightarrow \boldsymbol{R}_{m_1} \leftrightarrow ... \leftrightarrow \boldsymbol{R}_{m_s} \leftrightarrow \boldsymbol{R}_{b}
\end{equation}
where adjacent regions have nonempty intersections. Once equation (\ref{Equation 10: Connected convex regions}) is satisfied, GCS can establish edges between these convex regions and optimize a continuous trajectory. The remaining unprocessed path points are therefore no longer required.

To improve computational efficiency, multiple midpoint candidates generated according to the binary order are grouped into parallel batches. Each worker process executes the IRIS inflation algorithm using an independent robot context, and the temporary overlap graph is updated collectively after each batch. Since parallel inflation may generate multiple regions that contribute to connectivity within the same iteration, GCS-Bridging commits only the temporary regions belonging to the shortest overlapping chain once bridging is established. Regions generated in the current or previous batches that are not part of this shortest bridging chain are discarded. Therefore, additional candidates introduced by parallel inflation do not permanently increase the size of the GCS graph.

In summary, GCS-Bridging is a budget-constrained incremental reconnection algorithm rather than a complete decomposition method guaranteed to succeed in arbitrary free spaces. A feasible bridge may remain undiscovered within a given budget due to the restricted local sampling domain, finite random restarts, or limitations of convex region inflation. Nevertheless, the method preserves clear safety and consistency for all committed results. RRT candidate paths are densely collision-checked before use, each added convex region is generated by a collision-free region inflation algorithm, every GCS edge is explicitly verified by a nonempty region intersection, and intermediate results that do not form a complete overlapping chain are excluded from the global graph. These properties make GCS-Bridging particularly suitable as a topology-repair stage after initial convex decomposition, restoring the start-to-goal connectivity required for GCS planning while limiting the number of newly introduced regions.

\begin{figure}[t]
    \centering
    \includegraphics[width=0.48\textwidth]{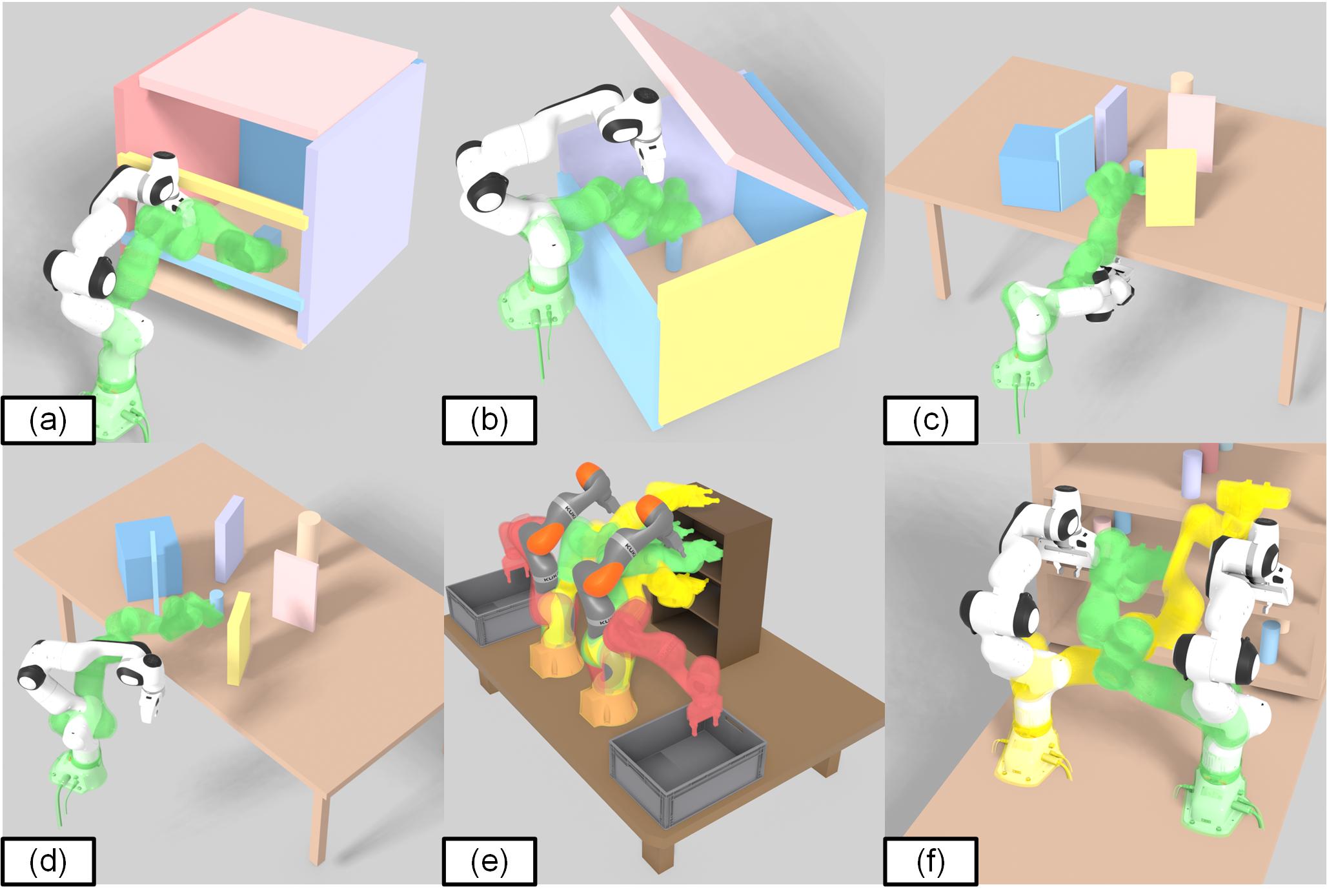}
    \caption{Multiple randomized test scenarios. Scenarios (a)–(d) are single-arm scenarios, while (e) and (f) are dual-arm scenarios. The single-arm scenarios are randomized using the framework in work \cite{chamzas2021motionbenchmaker}, with 100 trials conducted for each scenario. The dual-arm scenarios are randomized by us, with 50 trials conducted for each scenario. In (e), three colors indicate three distinct goals in the sequence solid $\rightarrow$ green $\rightarrow$ yellow $\rightarrow$ red. In (f), two colors distinguish the visually overlapping goals of the left and right arms.}
    \label{Figure 2: Multiple Scenarios Comparison}
\end{figure}

\section{Experimental Validation}
\label{Section IV}

\subsection{Methodology}
\label{Section IV.A}

The GCS-Bridging algorithm is evaluated across multiple IRIS algorithms and experimental scenarios. Specifically, (1) its compatibility with different IRIS algorithms is assessed by reconnecting the same GCS map using multiple IRIS inflation algorithms in an identical simulation scenario; (2) its performance under different GCS map reconnection strategies is analyzed across multiple simulation scenarios; and (3) a bookshelf point-to-point motion planning experiment is conducted on a physical Franka Emika Panda robot. All methods are implemented in compiled C++, with Python wrappers used when necessary. All tests are conducted on a PC equipped with an AMD Ryzen 9 7950X CPU (16-core, 3.6 GHz) and 128 GB of memory, while GPU-related computations are performed on an NVIDIA 5090 GPU.

\begin{table*}[t]
    \centering
    \caption{Random comparison of different GCS reconnection strategies in single-arm and dual-arm planning scenarios.}
    \label{Table: Multiple Reconnection Algorithm Comparison}
    \footnotesize
    \setlength{\tabcolsep}{4.5pt}
    \renewcommand{\arraystretch}{1.15}
    
    \begin{tabular}{
        c
        l
        l
        c
        c
        c
        c
        c
        c
    }
    \toprule
        \makecell[c]{\textbf{Robot} \\ \textbf{Type}}
        &
        \makecell[c]{\textbf{Scenario}}
        &
        \makecell[c]{\textbf{Reconnection} \\ \textbf{Method}}
        &
        \makecell[c]{$N^{\text{init}}_{\boldsymbol{R}}$ \\ $[\text{mean$\pm$std}]$}
        &
        \makecell[c]{$N^{\text{init}}_{\boldsymbol{V}}$ \\ $[\text{mean$\pm$std}]$}
        &
        \makecell[c]{$\Delta N_{\boldsymbol{R}}$ \\ $[\text{mean,[Q1,Q3]}]$}
        &
        \makecell[c]{\textbf{Success/}\\ \textbf{Trials}}
        &
        \makecell[c]{\textbf{Reconnection Time (s)} \\ $[\text{median,[Q1,Q3],P95}]$}
        &
        \\
        \midrule
        
        \multirow{12}{*}{\makecell{Single \\ Arm}}
        
        & \multirow{3}{*}{(a) Cage}
        & DirectLine
        & $\text{ }$
        & $\text{ }$
        & $\textbf{7.06,[5.00,9.00]}$
        & $\text{77/97}$
        & $\text{14.98,[11.73,18.57],25.80}$
        \\
        
        &
        & RRT-C
        & $\text{18.74} \pm \text{7.98}$
        & $\text{5.94} \pm \text{1.46}$
        & $\text{9.74,[7.00,11.00]}$
        & $\textbf{97/97}$
        & $\text{30.21,[14.56,37.43],42.47}$
        \\
        
        &
        & Hybrid
        & $\text{ }$
        & $\text{ }$
        & $\text{8.46,[6.00,10.00]}$
        & $\textbf{97/97}$
        & $\text{17.81,[14.23,24.11],76.24}$
        \\
        \cmidrule(lr){2-9}
        
        & \multirow{3}{*}{(b) Box}
        & DirectLine
        & $\text{ }$
        & $\text{ }$
        & $\textbf{2.93,[2.00,3.00]}$
        & $\text{93/100}$
        & $\text{21.31,[18.71,25.69],33.77}$
        \\
        
        &
        & RRT-C
        & $\text{15.21} \pm \text{6.17}$
        & $\text{5.94} \pm \text{1.53}$
        & $\text{3.30,[3.00,4.00]}$
        & $\textbf{100/100}$
        & $\text{10.20,[9.02,12.09],15.39}$
        \\
        
        &
        & Hybrid
        & $\text{ }$
        & $\text{ }$
        & $\text{3.05,[3.00,3.00]}$
        & $\textbf{100/100}$
        & $\text{21.66,[19.00,25.80],34.47}$
        \\
        \cmidrule(lr){2-9}
        
        & \multirow{3}{*}{(c) Table Under Pick}
        & DirectLine
        & $\text{ }$
        & $\text{ }$
        & $\textbf{2.48,[2.00,3.00]}$
        & $\text{83/100}$
        & $\text{12.33,[9.66,14.44],18.51}$
        \\
        
        &
        & RRT-C
        & $\text{12.88} \pm \text{5.43}$
        & $\text{5.86} \pm \text{1.36}$
        & $\text{2.83,[2.00,4.00}]$
        & $\textbf{100/100}$
        & $\text{9.64,[6.65,12.91],15.26}$
        \\
        
        &
        & Hybrid
        & $\text{ }$
        & $\text{ }$
        & $\text{2.73,[2.00,3.00}]$
        & $\textbf{100/100}$
        & $\text{12.55,[10.35,15.19],19.39}$
        \\
        \cmidrule(lr){2-9}

        & \multirow{3}{*}{(d) Table Pick}
        & DirectLine
        & $\text{ }$
        & $\text{ }$
        & $\textbf{1.98,[1.00,2.00}]$
        & $\text{91/100}$
        & $\text{8.04,[6.85,9.18],12.53}$
        \\
        
        &
        & RRT-C
        & $\text{12.88} \pm \text{5.43}$
        & $\text{5.86} \pm \text{1.36}$
        & $\text{2.19,[2.00,3.00]}$
        & $\textbf{100/100}$
        & $\text{5.00,[4.01,6.91],11.97}$
        \\
        
        &
        & Hybrid
        & $\text{ }$
        & $\text{ }$
        & $\text{2.09,[2.00,2.00]}$
        & $\textbf{100/100}$
        & $\text{8.19,[7.15,9.41],13.07}$
        \\
        
        \midrule
        
        \multirow{6}{*}{\makecell{Dual\\Arm}}
        
        & \multirow{3}{*}{(e) Dual IIWA}
        & DirectLine
        & $\text{ }$
        & $\text{ }$
        & $\text{4.18,[3.00,3.00]}$
        & $\text{0/50}$
        & $\text{63.49,[57.38,69.69],81.06}$
        \\
        
        &
        & RRT-C
        & $\text{21.46} \pm \text{2.21}$
        & $\text{6.00} \pm \text{0.00}$
        & $\text{11.30,[7.00,14.0]}$
        & $\textbf{50/50}$
        & $\text{778.78,[471.58,1219.09],2747.08}$
        \\
        
        &
        & Hybrid
        & $\text{ }$
        & $\text{ }$
        & $\textbf{10.56,[7.00,13.00]}$
        & $\textbf{50/50}$
        & $\text{741.99,[209.97,1514.07],3590.49}$
        \\
        \cmidrule(lr){2-9}
        
        & \multirow{3}{*}{(f) Dual Panda}
        & DirectLine
        & $ $
        & $ $
        & $\text{1.15,[1.00, 1.00]}$
        & $\text{0/50}$
        & $\text{717.00,[627.45,825.22],865.00}$
        \\
        
        &
        & RRT-C
        & $\text{10.70} \pm \text{3.74}$
        & $\text{3.52} \pm \text{0.52}$
        & $\text{16.70,[5.50, 26.50]}$
        & $\text{49/50}$
        & $\text{1660.44,[1051.11,3539.47],11939.32}$
        \\
        
        &
        & Hybrid
        & $ $
        & $ $
        & $\textbf{14.74,[5.00, 18.00]}$
        & $\textbf{50/50}$
        & $\text{1812.03,[1666.23,3182.89],16664.17}$
        \\
        
    \bottomrule
    \end{tabular}
    
    \vspace{1mm}
    \begin{minipage}{0.98\textwidth}
    \footnotesize{Column 2 specifies the test scenario. Column 3 identifies the three compared methods. Columns 4–8 report the experimental results, with the contents in square brackets is the representation of results. Column 4 gives the number of initially generated regions. Column 5 the initial number of connected components. Column 6 the number of additionally introduced regions. Column 7 the success rate of reconnecting the start and goal configurations, excluding initialization failures in which the start and goal configurations are already connected in the initial GCS map. Column 8 the reconnection time. Reconnection guarantees the existence of a connected region sequence, GCSTrajOpt optimization is not the focus of this comparison. The simulation results for the 6 scenarios are illustrated in \href{https://zhouxk1997.github.io/GCS_Bridging/result/GCS_Bridging_Multiple_Scenarios_Single_Arm_Cage.html}{result (a)}, \href{https://zhouxk1997.github.io/GCS_Bridging/result/GCS_Bridging_Multiple_Scenarios_Single_Arm_Box.html}{result (b)}, \href{https://zhouxk1997.github.io/GCS_Bridging/result/GCS_Bridging_Multiple_Scenarios_Single_Arm_Table_Under_Pick.html}{result (c)}, \href{https://zhouxk1997.github.io/GCS_Bridging/result/GCS_Bridging_Multiple_Scenarios_Single_Arm_Table_Pick.html}{result (d)}, \href{https://zhouxk1997.github.io/GCS_Bridging/result/GCS_Bridging_Multiple_Scenarios_Dual_Arm_IIWA.html}{result (e)}, and \href{https://zhouxk1997.github.io/GCS_Bridging/result/GCS_Bridging_Multiple_Scenarios_Dual_Arm_Franka.html}{result (f)}.}
    \end{minipage}
\end{table*}

\subsection{Multiple IRIS Algorithm Comparison}
\label{Section IV.B}
The KUKA bimanual scenario from \cite{marcucci2023motion} is adopted to evaluate the compatibility of GCS-Bridging with initial Graph of Convex Sets generated by four IRIS algorithms. In this scenario, 26 manually selected configurations are used as inflation seeds, and the resulting regions support sequential planning through 4 specified configurations. The evaluated algorithms include IRIS-NP\cite{marcucci2023motion}, IRIS-NP2\cite{werner2024faster}, IRIS-ZO\cite{werner2024faster}, and the CUDA-accelerated IRIS-ZO-CUDA developed with reference to \cite{werner2025superfast}, as the structure of IRIS-ZO is particularly suitable for CUDA acceleration. To ensure the quality of the generated regions, all IRIS algorithms were configured using high-fidelity parameter settings. In this experiment, 8 path points were inflated in parallel per iteration, and a CPU-based RRT-C algorithm was used for connection. Detailed results are reported in Table~\ref{Table: Multiple IRIS Algorithm Comparison}. This experiment consists of a single deterministic run and is intended for a qualitative comparison of compatibility across different IRIS algorithms. The trajectory cost in Table~\ref{Table: Multiple IRIS Algorithm Comparison} is defined by the total trajectory duration and the total path length in C-space as follows:
\begin{equation}
    \label{Equation 11: Trajectory cost}
    \text{Cost} = L(\boldsymbol{q}) + T
\end{equation}

GCS-Bridging demonstrates strong compatibility with all evaluated IRIS algorithms and successfully connects all start and goal configurations. This performance is primarily attributed to the suitability of RRT-C piecewise-linear paths for IRIS inflation and the ability of $\mathrm{LocalRRTC}$ to improve overall graph connectivity. Nevertheless, challenges remain. The computational cost of GCS-Bridging strongly depends on the connectivity of the initially generated graph. For IRIS-ZO and IRIS-ZO-CUDA, an excessive number of connected components in the initial GCS map substantially increases both the computational cost of $\mathrm{LocalRRTC}$ and $\mathrm{BinaryInflation}$ and the final optimization cost. This can be attributed to two factors. First, the faster IRIS-ZO and IRIS-ZO-CUDA algorithms generate relatively small convex regions, consistent with the comparative results reported in \cite{werner2024faster}. These small regions result in poor connectivity of the initial GCS map, requiring additional regions and producing a more fragmented feasible space. Consequently, the fully connected graph provides less flexibility for convex relaxation, leading to increased GCSTrajOpt cost. Second, because each $\mathrm{BinaryInflation}$ process inflates only 8 regions, unsuccessful connectivity after a single inflation batch requires repeated regions connectivity evaluations. The regions connectivity computation in IRIS involves numerous optimization problems, while each region contains a large number of facets, further increasing the Connectivity-Checking time, as particularly evident in the IRIS-ZO and IRIS-ZO-CUDA results.

\begin{figure*}[t]
    \centering
    \includegraphics[width=0.85\textwidth]{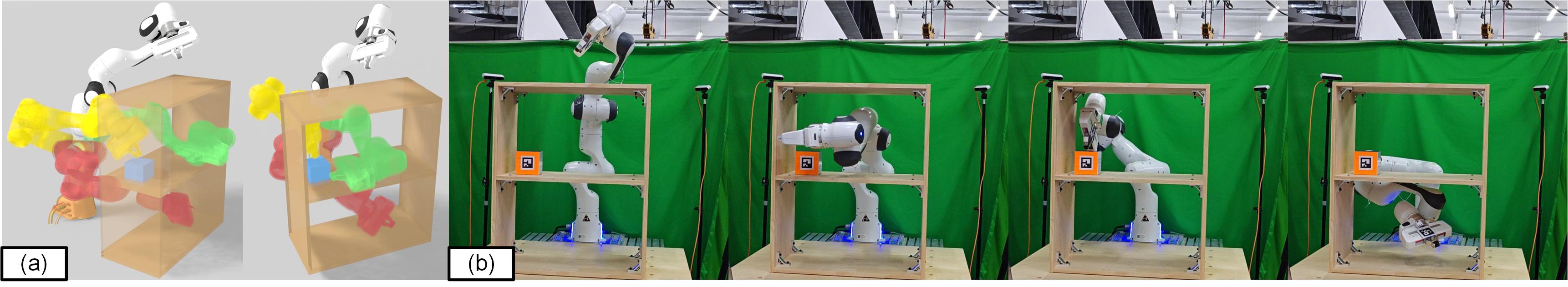}
    \caption{In the hardware experiment, the Franka Panda manipulator executes collision-free planning in the sequence solid $\to$ green $\to$ yellow $\to$ red shown in (a). The primary challenge is that the configurations represented by green and red are difficult to connect directly and therefore require reconnection through GCS-Bridging. The entire experimental execution in (b) remains collision-free. Simulation result is available \href{https://zhouxk1997.github.io/GCS_Bridging/result/GCS_Bridging_Physical_Experiment_Simulation.html}{here}, and hardware results are provided in \href{https://zhouxk1997.github.io/GCS_Bridging/result/GCS_Bridging_Physical_Experiment_View01.gif}{Video 1} and \href{https://zhouxk1997.github.io/GCS_Bridging/result/GCS_Bridging_Physical_Experiment_View02.gif}{Video 2}.}
    \label{Figure 3: Physical Experiment on Bookshelf Environment}
\end{figure*}

\subsection{Multiple Scenarios Comparison}
\label{Section IV.C}

As noted in \cite{werner2024approximating} and \cite{cohn2025non}, a GCS map with fewer, higher-quality regions generally yields faster solution times and better solution quality. Accordingly, seed points are often selected manually to obtain more favorable convex regions. When the initial GCS map cannot connect the start and goal configurations, manual reconnection strongly depends on environmental knowledge and prior experience, making systematic reproduction difficult. A simple alternative is $\mathrm{DirectLine}$ reconnection followed by inflation. By avoiding local path search, this strategy is computationally efficient. However, this $\mathrm{DirectLine}$ strategy becomes increasingly restrictive as environmental complexity and robot degrees of freedom increase, since the probability of collision-free linear connections between disconnected regions decreases in high-dimensional and constrained configuration spaces.

Based on these considerations, the following experiments compare $\mathrm{DirectLine}$ reconnection, RRT-C-based GCS-Bridging, and the hybrid $\mathrm{DirectLine}$/$\mathrm{LocalRRTC}$ strategy across multiple single-arm and dual-arm planning scenarios. The single-arm experiments use the cage, box, table under pick, and table pick scenarios from work \cite{chamzas2021motionbenchmaker}, while the dual-arm experiments use the iiwa bimanual scenario from work \cite{marcucci2023motion} and the dual-arm scenario from work \cite{moveit2} and \cite{moveit_benchmark_resources}.  All three methods are evaluated using identical initial convex-region decompositions, start and goal configurations. 100 repeated single-arm trials per scenario and 50 trials in dual-arm scenarios are conducted in each scenario on the same test instances to ensure paired comparisons. Performance is evaluated in terms of reconnection success rate, the number of additionally generated convex regions, and the total reconnection time. For large-scale randomized trials, the GPU-accelerated pRRTC algorithm\cite{huang2025prrtc} was adopted instead of the CPU-based RRT-C algorithm in single-arm scenarios, while dual-arm scenarios still use CPU-based RRT-C because pRRTC algorithm does not support centralized dual-arm planning. All tested scenarios are demonstrated in Fig.\ref{Figure 2: Multiple Scenarios Comparison}. The quantitative results are summarized in Table~\ref{Table: Multiple Reconnection Algorithm Comparison}.

Overall, across all scenarios, $\mathrm{DirectLine}$ achieves a success rate of 69.2\%, compared with 99.8\% for RRT-C-based method and 100\% for the hybrid strategy. Specifically, Table~\ref{Table: Multiple Reconnection Algorithm Comparison} shows that $\mathrm{DirectLine}$ achieves acceptable success rates in low-dimensional scenarios but degrades in geometrically complex or high-dimensional environments, such as Cage and the dual-arm scenarios. In contrast, RRT-C-based GCS-Bridging and the hybrid strategy achieve consistently higher success rates due to the effectiveness of RRT-C in high-dimensional C-space. Their overall performance is comparable in both success rate and computation time, while the shorter runtime of RRT-C-based method in some single-arm cases is primarily attributed to GPU acceleration. The hybrid strategy is therefore retained as a practical CPU-based option for low-dimensional scenarios. The substantially higher computation time observed for Dual Panda is mainly caused by its more complex collision model, which uses 66 spheres, compared with only 13 spheres for the simplified KUKA iiwa model, as shown in Fig.\ref{Figure 1: Overview of GCS-Bridging}(a).

\subsection{Hardware Validation on Bookshelf Environment}
\label{Section IV.D}

To evaluate the feasibility of GCS-Bridging in practical planning, we deploy it on a 7-DoF Franka Panda single-arm platform for point-to-point motion planning in a bookshelf environment. During the task, the manipulator must reach into the bookshelf while remaining collision-free throughout the motion. The environment is fully modeled using cameras before planning. The collision model of the Franka manipulator is simplified into an all-sphere representation using Foam\cite{coumar2025foam}, while the bookshelf is represented by cuboid collision models. The experimental setup is shown in Fig.~\ref{Figure 3: Physical Experiment on Bookshelf Environment}(a), and a snapshot of the hardware experiment is shown in Fig.~\ref{Figure 3: Physical Experiment on Bookshelf Environment}(b).

During the experiment, five convex regions containing the start, goals, and neutral configurations are first generated. Additional convex regions are then randomly initialized around the planning sequence, producing an initial GCS map in which the start-to-goal regions are not fully connected. In this scenario, the bookshelf partitions constitute particularly challenging obstacles for C-space convex regions inflation. Consequently, goal 1 (Fig.~\ref{Figure 3: Physical Experiment on Bookshelf Environment}(a) green) and goal 3 (Fig.~\ref{Figure 3: Physical Experiment on Bookshelf Environment}(a) red) in the initial GCS map are disconnected from other convex sets, making this environment well suited for evaluating GCS-Bridging. The experimental results in Fig.~\ref{Figure 3: Physical Experiment on Bookshelf Environment}(b) demonstrate that GCS-Bridging successfully reconnects the initially infeasible GCS map and generates feasible collision-free paths.

\section{Conclusion and Future Work}
\label{Section V}

In this work, we propose GCS-Bridging, a multi-region reconnection method for infeasible GCS planning problems based on joint-space Graph of Convex Sets construction and RRT-C. GCS-Bridging connects convex regions from different connected components in ascending order of their high-dimensional distances using either RRT-C or a hybrid strategy that applies $\mathrm{DirectLine}$ before $\mathrm{LocalRRTC}$, followed by convex region inflation. This enables GCS search and planning in configuration spaces with incomplete connectivity. The method is evaluated in an identical scenario initialized by different IRIS algorithms and in 6 randomized initially disconnected planning scenarios. Experimental results show that pure RRT-C-based GCS-Bridging achieves a 99.8\% success rate , while the hybrid strategy achieves 100\%. GCS-Bridging is further deployed on a 7-DoF single-arm robotic system, demonstrating that GCS-Bridging restores feasible GCS motion planning.

The current work is primarily limited by the IRIS-related convex region inflation algorithms in Drake. Existing collision-free convex region generation methods are either probabilistically collision-free IRIS-ZO \cite{werner2024faster}, EIZO\cite{werner2025superfast}, or based on deterministic heuristic guidance IRIS-NP\cite{marcucci2023motion}, IRIS-NP2\cite{werner2024faster}, and therefore do not provide complete collision-free guarantees. In contrast, methods with rigorous collision-free guarantees \cite{amice2022finding} suffer from the curse of dimensionality in high-dimensional spaces. Future work will focus on safer collision-free convex region generation in high-dimensional configuration spaces and the construction of more regularized convex regions.

\bibliographystyle{unsrt}
\bibliography{ZZZ_Main_Reference.bib}

@article{chen2026cssdf,
  title={CSSDF-Net: Safe Motion Planning Based on Neural Implicit Representations of Configuration Space Distance Field},
  author={Chen, Haohua and Zhou, Yixuan and Zhou, Yifan and Wang, Hesheng},
  journal={arXiv preprint arXiv:2603.18669},
  year={2026}
}

@article{li2024configuration,
  title={Configuration space distance fields for manipulation planning},
  author={Li, Yiming and Chi, Xuemin and Razmjoo, Amirreza and Calinon, Sylvain},
  journal={arXiv preprint arXiv:2406.01137},
  year={2024}
}

@article{lozano1983spatial,
  title={Spatial planning: A configuration space approach},
  author={Lozano-Perez, Tomas and others},
  journal={IEEE Trans. Computers},
  volume={32},
  number={2},
  pages={108--120},
  year={1983}
}

@inproceedings{thomason2024motions,
  title={Motions in microseconds via vectorized sampling-based planning},
  author={Thomason, Wil and Kingston, Zachary and Kavraki, Lydia E},
  booktitle={2024 IEEE international conference on robotics and automation (ICRA)},
  pages={8749--8756},
  year={2024},
  organization={IEEE}
}

@inproceedings{wilson2025nearest,
  title={Nearest-neighbourless asymptotically optimal motion planning with Fully Connected Informed Trees (FCIT*)},
  author={Wilson, Tyler S and Thomason, Wil and Kingston, Zachary and Kavraki, Lydia E and Gammell, Jonathan D},
  booktitle={2025 IEEE International Conference on Robotics and Automation (ICRA)},
  pages={14140--14146},
  year={2025},
  organization={IEEE}
}

@article{huang2025prrtc,
  title={prrtc: Gpu-parallel rrt-connect for fast, consistent, and low-cost motion planning},
  author={Huang, Chih H and Jadhav, Pranav and Plancher, Brian and Kingston, Zachary},
  journal={arXiv preprint arXiv:2503.06757},
  year={2025}
}

@article{sundaralingam2023curobo,
  title={curobo: Parallelized collision-free minimum-jerk robot motion generation},
  author={Sundaralingam, Balakumar and Hari, Siva Kumar Sastry and Fishman, Adam and Garrett, Caelan and Van Wyk, Karl and Blukis, Valts and Millane, Alexander and Oleynikova, Helen and Handa, Ankur and Ramos, Fabio and others},
  journal={arXiv preprint arXiv:2310.17274},
  year={2023}
}

@article{huang2024diffusionseeder,
  title={Diffusionseeder: Seeding motion optimization with diffusion for rapid motion planning},
  author={Huang, Huang and Sundaralingam, Balakumar and Mousavian, Arsalan and Murali, Adithyavairavan and Goldberg, Ken and Fox, Dieter},
  journal={arXiv preprint arXiv:2410.16727},
  year={2024}
}

@inproceedings{tam2024bomp,
  title={BOMP: Bin-Optimized Motion Planning},
  author={Tam, Zachary and Dharmarajan, Karthik and Qiu, Tianshuang and Avigal, Yahav and Ichnowski, Jeffrey and Goldberg, Ken},
  booktitle={2024 IEEE/RSJ International Conference on Intelligent Robots and Systems (IROS)},
  pages={11056--11063},
  year={2024},
  organization={IEEE}
}

@article{dalal2024neural,
  title={Neural mp: A generalist neural motion planner},
  author={Dalal, Murtaza and Yang, Jiahui and Mendonca, Russell and Khaky, Youssef and Salakhutdinov, Ruslan and Pathak, Deepak},
  journal={arXiv preprint arXiv:2409.05864},
  year={2024}
}

@article{zhang2024robotdiffuse,
  title={RobotDiffuse: Motion Planning for Redundant Manipulator based on Diffusion Model},
  author={Zhang, Xiaohan and Mou, Xudong and Wang, Rui and Wo, Tianyu and Gu, Ningbo and Wang, Tiejun and Xu, Cangbai and Liu, Xudong},
  journal={arXiv e-prints},
  pages={arXiv--2412},
  year={2024}
}

@article{yang2004sampling,
  title={The sampling-based neighborhood graph: An approach to computing and executing feedback motion strategies},
  author={Yang, Libo and Lavalle, Steven M},
  journal={IEEE Transactions on Robotics and Automation},
  volume={20},
  number={3},
  pages={419--432},
  year={2004},
  publisher={IEEE}
}

@inproceedings{deits2015computing,
  title={Computing large convex regions of obstacle-free space through semidefinite programming},
  author={Deits, Robin and Tedrake, Russ},
  booktitle={Algorithmic Foundations of Robotics XI: Selected Contributions of the Eleventh International Workshop on the Algorithmic Foundations of Robotics},
  pages={109--124},
  year={2015},
  organization={Springer}
}

@article{marcucci2023motion,
  title={Motion planning around obstacles with convex optimization},
  author={Marcucci, Tobia and Petersen, Mark and Von Wrangel, David and Tedrake, Russ},
  journal={Science robotics},
  volume={8},
  number={84},
  pages={eadf7843},
  year={2023},
  publisher={American Association for the Advancement of Science}
}

@article{werner2024faster,
  title={Faster algorithms for growing collision-free convex polytopes in robot configuration space},
  author={Werner, Peter and Cohn, Thomas and Jiang, Rebecca H and Seyde, Tim and Simchowitz, Max and Tedrake, Russ and Rus, Daniela},
  journal={The International Journal of Robotics Research},
  pages={02783649261436917},
  year={2024},
  publisher={SAGE Publications Sage UK: London, England}
}

@article{werner2025superfast,
  title={Superfast configuration-space convex set computation on GPUs for online motion planning},
  author={Werner, Peter and Cheng, Richard and Stewart, Tom and Tedrake, Russ and Rus, Daniela},
  journal={arXiv preprint arXiv:2504.10783},
  year={2025}
}

@article{clark2025plan,
  title={Plan Optimal Collision-Free Trajectories With Non-Convex Cost Functions Using Graphs of Convex Sets},
  author={Clark, Charles L and Xie, Biyun},
  journal={IEEE Transactions on Robotics},
  year={2025},
  publisher={IEEE}
}

@article{natarajan2024implicit,
  title={Implicit graph search for planning on graphs of convex sets},
  author={Natarajan, Ramkumar and Liu, Chaoqi and Choset, Howie and Likhachev, Maxim},
  journal={arXiv preprint arXiv:2410.08909},
  year={2024}
}

@inproceedings{amice2022finding,
  title={Finding and optimizing certified, collision-free regions in configuration space for robot manipulators},
  author={Amice, Alexandre and Dai, Hongkai and Werner, Peter and Zhang, Annan and Tedrake, Russ},
  booktitle={International Workshop on the Algorithmic Foundations of Robotics},
  pages={328--348},
  year={2022},
  organization={Springer}
}

@inproceedings{werner2024approximating,
  title={Approximating robot configuration spaces with few convex sets using clique covers of visibility graphs},
  author={Werner, Peter and Amice, Alexandre and Marcucci, Tobia and Rus, Daniela and Tedrake, Russ},
  booktitle={2024 IEEE International Conference on Robotics and Automation (ICRA)},
  pages={10359--10365},
  year={2024},
  organization={IEEE}
}

@article{cohn2025non,
  title={Non-Euclidean motion planning with graphs of geodesically convex sets},
  author={Cohn, Thomas and Petersen, Mark and Simchowitz, Max and Tedrake, Russ},
  journal={The International Journal of Robotics Research},
  volume={44},
  number={10-11},
  pages={1840--1862},
  year={2025},
  publisher={Sage Publications Sage UK: London, England}
}

@article{chamzas2021motionbenchmaker,
  title={Motionbenchmaker: A tool to generate and benchmark motion planning datasets},
  author={Chamzas, Constantinos and Quintero-Pena, Carlos and Kingston, Zachary and Orthey, Andreas and Rakita, Daniel and Gleicher, Michael and Toussaint, Marc and Kavraki, Lydia E},
  journal={IEEE Robotics and Automation Letters},
  volume={7},
  number={2},
  pages={882--889},
  year={2021},
  publisher={IEEE}
}

@article{coumar2025foam,
  title={Foam: A tool for spherical approximation of robot geometry},
  author={Coumar, Sai and Chang, Gilbert and Kodkani, Nihar and Kingston, Zachary},
  journal={arXiv preprint arXiv:2503.13704},
  year={2025}
}

@misc{moveit2,
  author       = {{MoveIt Maintainers and Core Contributors}},
  title        = {{MoveIt 2: The MoveIt Motion Planning Framework for ROS 2}},
  howpublished = {\url{https://github.com/moveit/moveit2}},
}

@misc{moveit_benchmark_resources,
  author       = {{Kayser, Henning and Jahr Sebastian  and Altiparmak, Cihat}},
  title        = {{MoveIt Benchmark Resources}},
  howpublished = {\url{https://github.com/moveit/moveit_benchmark_resources}},
}

\end{document}